\documentclass[a4paper, 11pt, oneside]{article} 

\usepackage{geometry}
\usepackage{booktabs} 

\usepackage{mathptmx} 

\usepackage{fancyhdr}
\usepackage[normalem]{ulem}
\usepackage{microtype}
\usepackage{graphicx}
\usepackage{subfigure}
\usepackage{colortbl}
\definecolor{mygray}{gray}{.88}
\usepackage{multirow}
\usepackage{amsmath}
\usepackage{bm}
\usepackage{epsfig}
\usepackage{authblk}

\usepackage[hyphens]{url}
\usepackage{hyperref}
\usepackage{color}
\usepackage{soul}
\usepackage{bbding}

\usepackage[T1]{fontenc}
\usepackage{graphicx}
\usepackage{hyperref}       
\usepackage{url}            
\usepackage{booktabs}       
\usepackage{amsfonts}       
\usepackage{algorithm}
\usepackage{algpseudocode}
\usepackage{multirow}
\usepackage{tabularx}

\usepackage{graphicx}
\usepackage[normalem]{ulem}
\useunder{\uline}{\ul}{}

\usepackage[bottom]{footmisc} 
\usepackage{CJK}

\begin{document} 
\begin{CJK}{UTF8}{gbsn}
\begin{titlepage} 

	\centering 
	
	\scshape 
	
	\vspace*{\baselineskip} 
	
	
	\rule{\textwidth}{1.6pt}\vspace*{-\baselineskip}\vspace*{2pt} 
	\rule{\textwidth}{0.4pt} 
	
	\vspace{0.75\baselineskip} 
	
	{\LARGE AGIBench: A Multi-granularity, Multimodal, Human-referenced, Auto-scoring Benchmark for Large Language Models\\} 
	
	\vspace{0.75\baselineskip} 
	
	\rule{\textwidth}{0.4pt}\vspace*{-\baselineskip}\vspace{3.2pt} 
	\rule{\textwidth}{1.6pt} 
	
	\vspace{2\baselineskip} 
	
	
	
	\vspace*{3\baselineskip} 
	
	
	Edited By
	
	\vspace{0.5\baselineskip} 
	
	{\scshape\Large  Fei Tang\\ Wanling Gao\\ Luzhou Peng\\Jianfeng Zhan\\ }
	
	
	\vspace{0.5\baselineskip} 

	\vfill 
	
	
	\epsfig{file=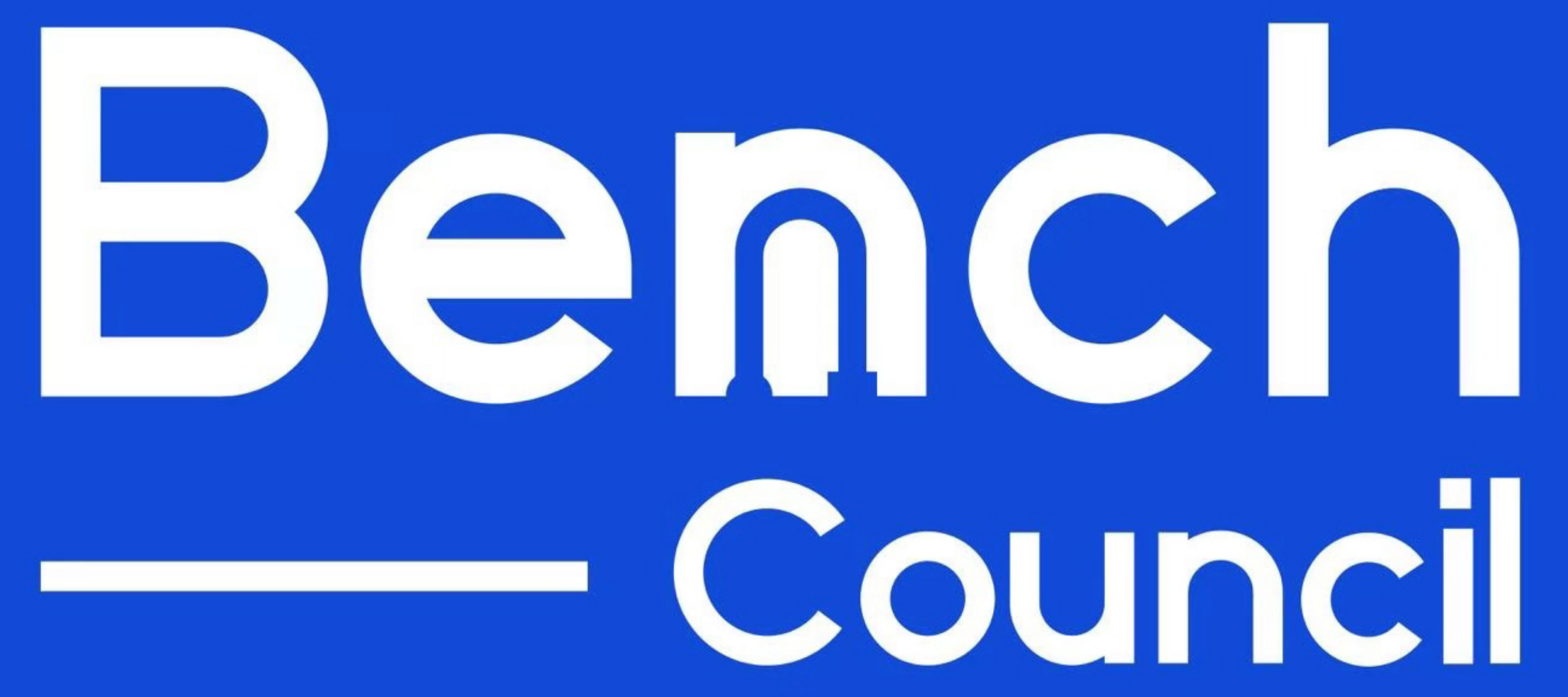,height=2cm}
	\textit{\\BenchCouncil: International Open Benchmark Council\\Chinese Academy of Sciences\\Beijing, China\\http://www.benchcouncil.org/agibench} 
	\vspace{5\baselineskip} 

	Technical Report No. BenchCouncil-AGIBench-2023 
	
	{\large Aug 22, 2023} 

\end{titlepage}


\title{AGIBench: A Multi-granularity, Multimodal, Human-referenced, Auto-scoring Benchmark for Large Language Models}

\author[1,3]{Fei Tang}
\author[1,2,3]{Wanling Gao}
\author[4]{Luzhou Peng}
\author[1,2,3]{Jianfeng Zhan\thanks{Jianfeng Zhan is the corresponding author.}}

\affil[1]{Research Center for Advanced Computer Systems, State Key Lab of Processors, Institute of Computing Technology, Chinese Academy of Sciences\\ \{tangfei, gaowanling,  zhanjianfeng\}@ict.ac.cn}
\affil[2]{BenchCouncil (International Open Benchmark Council)}
\affil[3]{University of Chinese Academy of Sciences}
\affil[4]{Shanghai Lixin University of Accounting and Finance}

\date{Aug 22, 2023}
\maketitle

\begin{abstract}
Large language models (LLMs) like ChatGPT have revealed amazing intelligence. How to evaluate the question-solving abilities of LLMs and their degrees of intelligence is a hot-spot but challenging issue.
First, the question-solving abilities are interlaced with different ability branches like understanding and massive knowledge categories like mathematics. 
Second, the inputs of questions are multimodal that may involve text and images. In addition, they may have varying levels of difficulty while lacking a unified standard to judge which one is more difficult.
Third, the response format of LLMs is diverse and thus poses great challenges for result extraction and evaluation.
Even though several benchmarks have been proposed to evaluate LLMs, however, a majority of them only evaluate and report the performance on a collection of blended text questions that may
 involve multiple ability branches, knowledge categories, or different difficulty levels. Thus, the evaluation may result in indistinguishable performance data and a biased optimization direction. A small fraction of efforts attempt to classify the questions according to the ability branches, however, they still provide coarse-grained benchmarking such as lacking detailed knowledge categories and human-referenced difficulty levels. In addition, existing efforts either adopt prompt engineering to normalize the response format or analyze the result manually, which would introduce unpredictable performance impacts or unacceptable evaluation costs.

In this paper, to tackle the above challenges, we propose AGIBench -- a multi-granularity, multimodal, human-referenced, and auto-scoring benchmarking methodology for LLMs. Instead of a collection of blended questions, AGIBench focuses on three typical ability branches and adopts a four-tuple <ability branch, knowledge, difficulty, modal> to label the attributes of each question. 
First, it supports multi-granularity benchmarking, e.g., per-question, per-ability branch, per-knowledge, per-modal, per-dataset, and per-difficulty level granularities. Second, it contains multimodal input, including text and images. Third, it classifies all the questions into five degrees of difficulty according to the average accuracy rate of abundant educated humans (human-referenced). Fourth, it adopts zero-shot learning to avoid introducing additional unpredictability and provides an auto-scoring method to extract and judge the result. Finally, it defines multi-dimensional metrics, including accuracy under the average, worst, best, and majority voting cases, and repeatability. 
Our experiments on twelve state-of-the-art LLMs show the effectiveness of our benchmark. AGIBench is publically available from \url{https://www.benchcouncil.org/agibench}.


\end{abstract}


\section{Introduction}

Intelligence is an abstract concept and has no unified definition yet~\cite{legg2007collection}. Research about human intelligence has been conducted for decades and formed a series of theories, e.g., triarchic theory of intelligence~\cite{sternberg1997triarchic}, fluid and crystallized intelligence~\cite{cattell1963theory}, theory of multiple intelligences~\cite{gardner1987theory}, etc., while having no unified standard about how to evaluate human intelligence. 
In this condition, the difficulties aggravate evaluating artificial intelligence (AI), which has shown powerful abilities to solve problems or questions and reflects the tremendous potential to approach human intelligence, especially the emerging large language models (LLMs) like ChatGPT. 
Different from the previous AI applications that mainly target a single task or a specific domain, LLMs anticipate achieving general intelligence. Hence, the previous benchmarking methodologies that focus on specific tasks or application domains are not applicable anymore. A new benchmarking methodology is a necessity but no easy feat.

On the one hand, how to construct a benchmark with diverse, typical, and difficulty-differentiated input questions is challenging. Similar to human intelligence, the intelligence of LLMs has no unified benchmarking standard since the input questions or problems to be solved are massive and interlaced. First, the input questions or problems may not only involve multiple ability branches like understanding and reasoning but also involve numerous knowledge categories with varying levels of difficulties like mathematics and geography.
Second, the input questions or problems are multimodal with a combination of text and images, for example, geometry problems in mathematics, talking about pictures in linguistics, etc.

On the other hand, how to evaluate the benchmarking results and choose comprehensive and important metrics are challenging. From the perspective of analyzing the results, the response formats of LLMs are diverse. For example, the response may (1) answer the choice from four choices marked A, B, C, and D, with an explanation and analysis of the above four choices one by one; (2) only answer the choice without an explanation; (3) answer the choice with a partial explanation. In addition, the orders of choices and explanations vary, and the text words are discrepant. Even though a human being can distinguish the result easily, however, it is extremely hard for a computer program. From the perspective of the evaluation metrics, the performance of LLMs is multidimensional and not merely average accuracy. For example, the LLMs may be adept in specific ability branches, knowledge categories, or difficulty levels; the response results may change during multiple runs, etc. 

\newcolumntype{Y}{>{\hsize=0.5\hsize}X}
\newcolumntype{Z}{>{\hsize=0.5\hsize}X}

Many efforts have been proposed to benchmark the LLMs~\cite{srivastava2023beyond,liang2022holistic,zheng2023judging,zhong2023agieval,huang2023ceval,zhang2023evaluating}; however, they fail to solve the above challenges. 
First, almost all adopt one or more open-source datasets with a collection of blended text questions at a per-dataset or per-ability branch granularity. However, these blended questions may cover different knowledge categories and different difficulty levels. We may hardly distinguish the performance on each ability branch, knowledge, or difficulty level, except for a total score on a dataset. Furthermore, even some of those benchmarks~\cite{zhong2023agieval} provide different difficulty levels; however, these difficulty levels are applied to a per-dataset granularity, which means the overall difficulty of the entire question dataset while not a specific question. 
Second, to solve the challenges of analyzing the response results, most of them~\cite{liang2022holistic,zhong2023agieval,huang2023ceval} adopt prompt engineering like few-shot and chain-of-thought (CoT) to increase the accuracy or normalize the response format. However, our experiments and the related work show that no matter few-shot or CoT would introduce unpredictable performance impacts. Our experiments especially show that with few-shot, the accuracy variance achieves to 5\% when using different random seeds. C-Eval~\cite{huang2023ceval} shows that with CoT, the accuracy increases on some models (e.g., ChatGLM-6B, +3\%) while decreases on other models (e.g., Chinese-LLaMA-13B, -11.9\%). Some of the efforts~\cite{zhang2023evaluating} adopt a zero-shot approach to avoid the unpredictable impact, however, they have to analyze the response result manually, which results in unacceptable evaluation costs.
Third, most of those efforts only report average accuracy and lack a comprehensive and multidimensional evaluation. While HELM~\cite{liang2022holistic} incorporates multiple metrics, it omits human-referenced evaluation and does not include a multimodal dataset.

\begin{table}[H]
\centering
\scriptsize
\caption{Observations and Implications for LLMs using AGIBench.}
\label{tab:insights}
\begin{tabular}{|p{3.5in}|p{1.2in}|}
\hline
\textbf{Observations} & \textbf{Implications} \\ 
\hline
\textbf{Multi-granularity:} (1) ChatGLM v2-6B outperforms ChatGLM-130B even with less parameters; 
(2) LLMs reflect good common sense (e.g., GPT-4 outperforms humans by 15.84\%) and understanding ability (e.g., GPT-4 is comparable with humans) while poor reasoning ability (e.g., GPT-4 underperforms humans by 25.94\%). 
(3) GPT-4 has the highest performance on most ability branches and knowledge categories. 
& (1) Architecture improvement and high-quality training data are more pivotal than large model size; (2) Reasoning ability is a direction of optimization. \\ 
\hline
\textbf{Multimodal:} (1) The image understanding and reasoning abilities are poor for the four LLMs that support open image access. (2) For LLMs that have no image processing ability, a majority of the responses cannot admit the limitation and output hallucinational and nonsense responses.
& (1) The multimodal abilities need optimizations. \\
\hline
\textbf{Human-referenced Difficulty:}  (1) Humans perform better than LLMs on simple questions (i.e., Level 1 to 3) while worse on difficult questions (i.e., Level 4 and 5). 
(2) The accuracy of GPT-4 is higher than humans for common sense, comparable for understanding, and worse for reasoning. 
& (1) The solving abilities of simple questions and reasoning need optimizations. \\
\hline
\textbf{Multi-dimensional metrics:}  
(1) The worst-case accuracy of LLMs is significantly below the corresponding average one, which means the model does not always give a correct answer during three times evaluations, indicating a poor reliability of LLMs. (2) The best-case accuracy of LLMs is much higher than the other cases, which means the model has a high probability to give a correct answer during three evaluations. (3) The majority voting accuracy is similar to the average one, which means most of the time, the model can give a correct answer.
& 
(1) The reliability of LLMs needs optimization. \\
\hline
\end{tabular}%
\end{table}

This paper proposes AGIBench -- a multi-granularity, multimodal, human-referenced, and auto-scoring benchmarking methodology and benchmark for LLMs. Instead of a collection of blended questions, AGIBench focuses on fundamental ability branches and adopts a four-tuple <ability branch, knowledge, difficulty, modal> to label the attributes of each question. 
In total, AGIBench provides 927 questions, covering three kinds of ability branches, i.e., common sense, reasoning, and understanding, 20 knowledge categories and 68 knowledge subclasses.
First, it supports multi-granularity benchmarking, e.g., per-question, per-ability branch, per-knowledge, per-difficulty level, per-dataset, and per-modal granularities. Second, it contains multimodal input, including various contexts like text-only, image-only, text with images, text with tables, and a combination of text, images, and tables. Third, it classifies all the questions into five degrees of difficulty according to the average accuracy rate of abundant educated humans (human-referenced). Fourth, it adopts zero-shot learning to avoid additional unpredictability and provides an auto-scoring method to extract and judge the result.  Finally, it defines multi-dimensional metrics, including accuracy under the average, worst, best, and majority voting cases, and repeatability. 
Our experiments on twelve state-of-the-art LLMs show the effectiveness of our benchmark. Table~\ref{tab:insights} presents the main observations and implications of LLMs using AGIBench.


\section{Related Work}

Many efforts have been proposed to evaluate traditional natural language processing (NLP) algorithms and LLMs.
GLUE~\cite{wang2018glue} and SuperGLUE~\cite{wang2019superglue} are benchmarks for traditional NLP tasks, primarily evaluating understanding capabilities. However, the questions are limited in scope and mainly focus on sentence classification, which cannot comprehensively reflect the complexities of human language~\cite{raji2021ai}. Thus, these benchmarks cannot meet the need for evaluating LLMs.

For evaluating LLMs, from the perspective of the question dataset, most of the related work uses closed-ended questions since the answer is definitive without subjectivity. BIG-Bench~\cite{srivastava2023beyond} and HELM~\cite{liang2022holistic} use a combination of multiple datasets like MATH~\cite{hendrycks2021measuring} and GSM8K~\cite{cobbe2021training}. On the one hand, such an approach may incur question redundancy and result in limited coverage. On the other hand, they only support the benchmarking at the per-dataset or per-ability branch granularity and thus cannot reveal the abilities from different perspectives. 
For example, they only output a score on a specific dataset with blended questions that involve different difficulty levels. 
Hence, we can hardly know the ability for a specific knowledge category or a specific difficulty level.  
ScienceQA~\cite{raji2021ai} collects questions from elementary and high school science curricula. However, these questions are too easy for humans. AGI-Eval~\cite{zhong2023agieval}, C-Eval~\cite{huang2023ceval}, and GAOKAO~\cite{zhang2023evaluating} focus on the LLMs benchmarking using the Chinese language. They utilize exams like national civil service and college entrance exams as question datasets. However, these benchmarks only use average accuracy as the evaluation metric, which is overly simplistic.
Several efforts attempt to use open-ended questions for LLMs benchmarking.
Chandrasekaran et al. ~\cite{bubeck2023sparks} collect a series of open-ended and multimodal questions to evaluate GPT-4; however, evaluating and scoring the results is extremely hard. In addition, the evaluation results are hard to reproduce, considering the subjectivity of different persons.




From the perspective of evaluation and scoring method, a majority of the existing efforts adopt prompt engineering for evaluation, such as few-shot learning~\cite{wang2020generalizing} and Chain-of-Thought (CoT)~\cite{wei2022chain}. Although these methods have been proven to be the potential to standardize the format of the response result or increase model accuracy, however, C-Eval~\cite{huang2023ceval} and our experiments show that they may not always be effective and would introduce unpredictable performance impacts. In this condition, we cannot evaluate the abilities of LLMs accurately since the response results may be impacted by prompt engineering. Additionally, using these methods requires case-by-case tuning, which exacerbates the benchmarking costs and cannot assure the fairness of benchmarking. 
MT-Bench~\cite{zheng2023judging} explores the use of GPT-4 as an approach to judge the correctness of the response results of LLMs and compare it with human judgment. They find that the decisions from GPT-4 and humans have an 80\% similarity. However, MT-Bench only selects 80 questions and thus has limited representativeness. Additionally, using GPT-4 directly for scoring would incur huge labor costs since we still need to check the judgment for every question.



\section{The Design and Implementation}

This section illustrates the design and implementation of AGIBench. Section~\ref{methodology} describes the benchmarking methodology. Section~\ref{design} presents an AGIBench overview. 

\subsection{Methodology}~\label{methodology}

To evaluate the different question-solving abilities of LLMs and their degrees of intelligence, we adopt a multi-granularity, multimodal, human-referenced, and auto-scoring benchmarking methodology, covering the question dataset construction, evaluation methodology, and evaluation metrics. 

First, to support multi-granularity benchmarking instead of only per-dataset or per-ability branch benchmarking adopted in the related work, we use a four-tuple <ability branch, knowledge, difficulty, modal> to label the attributes of each question. The ability branch focuses on the most fundamental and essential abilities like understanding. Knowledge includes a broad spectrum of knowledge categories within each ability branch, e.g., passage reading within understanding ability. Difficulty indicates a question's difficulty level, i.e., Levels 1 to 5, from easy to difficult. Modal indicates the modal of a question like text or image. 

Second, to construct a representative and typical question dataset, we aim to cover many questions with diverse and varied attributes. Specifically, we choose the three most fundamental ability branches: common sense, understanding, and reasoning. We single out twenty comprehensive and representative knowledge categories for these three ability branches in total. Knowledge for common sense covers six categories: humanities, technology, law, geography, politics, and economy; Knowledge for understanding includes passage reading, sentence grammar, fill-in-the-blank, and long text reading;
Knowledge for reasoning includes graphical reasoning, definition judgment, comprehensive materials, tabular materials, textual materials, graphical materials, analogical reasoning, logical judgment, mathematical calculation, and numerical reasoning.
For the difficulty attribute of each question, we adopt a human-referenced methodology based on big data statistics, which uses the accuracy rate answered by millions of well-educated humans to label the question's difficulty level. For example, Level 1 is the easiest one with an 80\% to 100\% accuracy rate, which means 80\% to 100\% of millions of humans can give a correct answer. Level 2 has a 60\% to 80\% accuracy rate. Level 3 has a 40\% to 60\% accuracy rate. Level 4 has a 20\% to 40\% accuracy rate. Level 5 is the most difficult and has a 0\% to 20\% accuracy rate.
We cover multimodal input that encompasses various contexts, including text-only, text with tables, text with images, image-only, and a combination of text, images, and tables.

Third, to cope with the diversity of the response format and avoid the unpredictable performance impacts of prompt engineering, we do not use prompt engineering and attempt to find a series of regex patterns to extract the answers and perform judgment.

Fourth, we define multi-dimensional metrics for LLMs benchmarking: accuracy under different cases and repeatability. 





\subsection{AGIBench Design and Implementation}~\label{design}

We design and implement the AGIBench based on the methodology, including the question dataset construction, evaluation and scoring, and metrics.

\begin{figure}[h]
    \centering
    \includegraphics[width=0.95\textwidth]{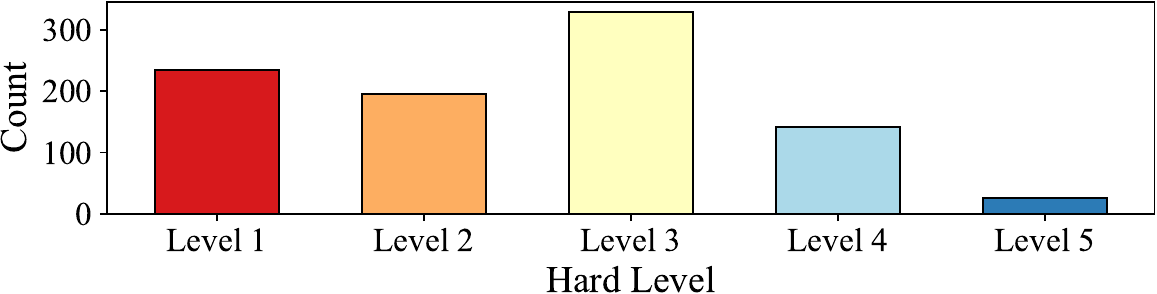}
    \caption{The Distribution of Five Difficulty Levels. Using the accuracy rate answered by millions of well-educated humans as references.}
    \label{fig:human_accuracy}
\end{figure}

\textbf{Question Dataset.}
AGIBench selects questions related to human life, especially for the Chinese. We mainly choose the questions from national civil service examinations since they satisfy the diversity and fundamentality requirements. 
We use a four-tuple <ability branch,
knowledge, difficulty, modal> to label the attributes of each question.
The ability branch covers common sense, understanding, and reasoning abilities. The knowledge contains 20 categories and 68 subclasses covering humanities, physics, chemistry, economics, law, politics, culture, geography, history, engineering, mathematics, etc. 
Tab.~\ref{tab:agibench_dataset} shows the ability branch and knowledge attributes of the AGIBench dataset. 
Regarding difficulty, our dataset uses human accuracy as the reference, and the accuracy is based on the high-educated human. 
We carefully select the difficulty of questions and classify five levels from Levels 1 to Level 5. A higher number means a more challenging degree. From Level 1 to Level 5, the responding human accuracy is [80\%, 100\%], [60\%, 80\%), [40\%, 60\%), [20\%, 40\%), and [0\%, 20\%). Fig.~\ref{fig:human_accuracy} shows the distribution of difficulty levels of AGIBench.
As for the modal, 
we include multimodal input covering various contexts, i.e., text-only, image-only, text with images, text with tables, and a combination of text, images, and tables. The corresponding number of questions for these contexts are 863, 5, 38, 10, and 11, respectively. In total, we provides 927 questions. Note that the image is processed as a URL in the questions.
Additionally, the text dataset contains plain text, complex mathematical formulas, and table data. We adopt the latex format for mathematical formulas, and for table data, we use the markdown format.


\begin{table}[H]
\centering
\caption{Ability Branch and Knowledge Attributes of Question Dataset in AGIBench.}
\label{tab:agibench_dataset}
\resizebox{\textwidth}{!}{%
\begin{tabular}{@{}llp{12cm}@{}}
\toprule
Ability Branch & Knowledge & Knowledge Subclass (Percentage) \\ \midrule
\multirow{6}{*}{Common Sense} & Economics & Economics (1.08\%) \\ \cmidrule(l){2-3} 
 & Geography & Environmental (0.76\%), National and Social Conditions (1.08\%), Natural   (1.08\%) \\ \cmidrule(l){2-3} 
 & Humanities & Chinese History (1.08\%), Cultural (1.08\%), Literary (1.08\%), World   History (1.08\%) \\ \cmidrule(l){2-3} 
 & Law & Administrative Law (1.08\%), Civil Law (1.08\%), Commercial and Economic   Law (1.08\%), Constitutional Law (1.08\%), Criminal Law (0.97\%), Jurisprudence   (1.08\%), Procedural Law (2.16\%) \\ \cmidrule(l){2-3} 
 & Politics & Politics (1.08\%) \\ \cmidrule(l){2-3} 
 & Technology & Biology Fundamentals (1.08\%), Chemistry Fundamentals (1.08\%), Everyday   Knowledge (1.08\%), Physics Fundamentals (1.08\%), Technology Theories and   Achievements (1.08\%) \\ \midrule
\multirow{10}{*}{Reasoning} & Analogical Reasoning & Grammatical Relations (0.86\%), Logical Relations (5.07\%), Semantic   Relations (3.02\%) \\ \cmidrule(l){2-3} 
 & Comprehensive Materials & Comprehensive Materials (1.08\%) \\ \cmidrule(l){2-3} 
 & Definition Judgment & Multiple Definitions (1.08\%), Single Definition (1.08\%) \\ \cmidrule(l){2-3} 
 & Graphic Materials & Graphic Materials (1.08\%) \\ \cmidrule(l){2-3} 
 & Graphic Reasoning & Attribute Principles (0.86\%), Pattern Principles (1.83\%), Positional   Principles (1.94\%), Quantity Principles (1.4\%), Spatial Reconstruction   (0.22\%), Special Principles (0.86\%) \\ \cmidrule(l){2-3} 
 & Logical Judgment & Combinations and Arrangements (1.08\%), Daily Conclusions (1.08\%), Reason   Explanations (0.97\%), Strengthened Types (1.08\%), Translation Reasoning   (1.08\%), True-False Reasoning (0.97\%), Weakened Types (1.08\%) \\ \cmidrule(l){2-3} 
 & Mathematical Calculation & Core Methods (5.18\%), Economic Profit and Comprehensive Planning (2.91\%),   Engineering (1.08\%), Inclusion-Exclusion Principle (12.51\%), Journey (1.08\%),   Permutation, Combination, and Probability (2.16\%), Solution Problems (0.97\%) \\ \cmidrule(l){2-3} 
 & Numerical Reasoning & Basic Sequences (0.76\%), Exponential Sequences (0.97\%), Fractional   Sequences (1.08\%), Mechanical Split Sequences (1.94\%), Multi-level Sequences   (1.08\%), Multiple Sequences (0.54\%), Recursive Sequences (1.08\%) \\ \cmidrule(l){2-3} 
 & Tabular Materials & Tabular Materials (1.08\%) \\ \cmidrule(l){2-3} 
 & Textual Materials & Textual Materials (1.08\%) \\ \midrule
\multirow{4}{*}{Understanding} & Fill-in-the-blank & Content Word Fill-in-the-blank (1.08\%), Idiom Fill-in-the-blank (1.08\%),   Mixed Fill-in-the-blank (1.08\%) \\ \cmidrule(l){2-3} 
 & Long Text Reading & Long Text Reading (1.08\%) \\ \cmidrule(l){2-3} 
 & Passage Reading & Central Understanding (1.62\%), Detail Judgment (1.08\%), Sentence   Understanding (1.08\%), Title Insertion (1.08\%) \\ \cmidrule(l){2-3} 
 & Sentence Grammar & Flawed and Ambiguous (1.08\%), Following Sentence Choice (1.08\%), Sentence   Fill-in-the-blank (1.08\%), Sentence Ordering (1.08\%) \\ \bottomrule
\end{tabular}%
}
\end{table}

\begin{figure}[h]
    \centering
    \includegraphics[width=0.95\textwidth]{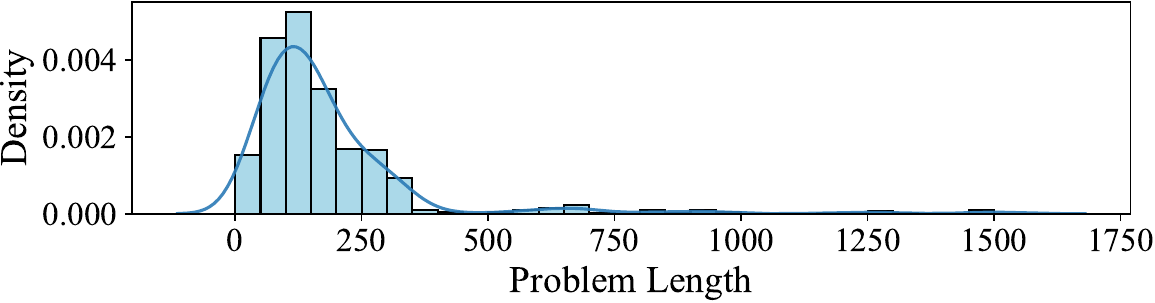}
    \caption{The Length Distribution of the Question Dataset in AGIBench.}
    \label{fig:problem_length}
\end{figure}

We further consider the length of questions. Fig~\ref{fig:problem_length} shows the length distribution of questions, and our dataset covers a broad spectrum. The size of most questions is more significant than 100 while many others are less than ten words~\cite{lu2022learn}. And our dataset contains several long-length questions whose length exceeds 1000, which also shows the difficulty and variety of our dataset. 

\textbf{Evaluation and Scoring.}
To avoid the impact of prompt engineering and meanwhile essentially reduce labor costs, we adopt an auto-scoring methodology that combines a heuristic regular expression searching algorithm (HRE for short) and GPT-4.  
On the one hand, we use an HRE algorithm to search the regex patterns, as shown in Algorithm~\ref{alg:pattern_sets}. We repeat N iterations and randomly select M responses from the total responses during each iteration. For the M responses of each iteration, we set a threshold ``minimum\_limit", indicating the minimum number of occurrences that a response format can be added as a new regex pattern. After that, we obtain a set that contains frequently occurring regex patterns. 
Our evaluation statistics in Section~\ref{evasetup} verify the effectiveness of the HRE algorithm. Among hundreds of thousands of responses from LLMs, about 67\% of the response results can be adequately extracted using HRE.
On the other hand, in terms of the remaining small fraction that cannot be extracted by HRE, we use GPT-4 to extract the results.
Note that we do not use GPT-4 directly because its accuracy cannot achieve 100\%, and we still need to verify the judgments artificially. By adopting HRE, we reduce about seventy percent labor costs.

\begin{algorithm}[ht]
\caption{Heuristic regular expression searching (HRE)}
\label{alg:pattern_sets}
\begin{algorithmic}[1]
\State pattern\_sets $\gets$ []
\For{$i \gets 1$ \textbf{to} $N$}
    \State sampled\_responses $\gets$ sample(total\_responses, M)
    \While{max\_number\_of\_common\_patterns(sampled\_responses) $>$ minimum\_limit}
        \State temp\_pattern $\gets$ propose\_pattern(sampled\_responses)
        \State patched\_to\_pattern\_sets $\gets$ False
        \For{pattern $\in$ pattern\_sets}
            \If{can\_patch(pattern, temp\_pattern)}
                \State pattern $\gets$ patch(pattern, temp\_pattern)
                \State patched\_to\_pattern\_sets $\gets$ True
                \State \textbf{break}
            \EndIf
        \EndFor
        \If{not patched\_to\_pattern\_sets}
            \State pattern\_sets.append(temp\_pattern)
        \EndIf
        \State sampled\_responses $\gets$ sampled\_responses $-$ match(sampled\_responses, temp\_pattern)
    \EndWhile
\EndFor
\end{algorithmic}
\end{algorithm}

\textbf{Metrics.}
We collect many metrics to evaluate LLMs comprehensively, not merely average accuracy, which is the only metric for most related work. We widely include the average accuracy, the worst-case accuracy, the best-case accuracy, the majority voting accuracy, and the repeatability to indicate the performance of LLMs under different cases. 
We detail these metrics as follows. Note that 
we evaluate each LLM using a question for three times to assure the fairness and the reliability of benchmarking. For each time, if the answer is correct, the score is 1, otherwise 0.

(1) Average accuracy. For the three times' evaluations on an LLM using the same question, we use the average score as the score of the LLM for that question. Then we calculate the average score for all questions as the final average accuracy.

(2) The worst-case accuracy. Different from the average accuracy, if all three times' evaluations give the correct answer, the score is 1, otherwise 0. Then the average value on all questions is reported as the final worst-case accuracy.

(3) The best-case accuracy. More relaxed compared to the worst-case one, if great than or equal to one answer gives the correct answer, the score is 1, otherwise 0. 

(4) The majority voting accuracy. If at least two answers are correct, the score is 1, otherwise 0.


(5) Repeatability. The similarity of the responses during three different runs. A high similarity indicates good repeatability.

\section{Evaluation}
This section presents the evaluation, including the evaluation methodology (Section~\ref{evameth}), experiment setup~\ref{evasetup}, and evaluation results~\ref{evaresult}.

\subsection{Evaluation Methodology}~\label{evameth}

\subsubsection{LLMs Overview}
We choose representative LLMs with different underlying technology, differentiated model sizes, and state-of-the-art performance. We also consider additional training data, architectures, target purposes, and whether open sourced. 
Specifically, we choose 12 models from OpenAI, Anthropic, Meta, Tsinghua, Baidu, Alibaba, and iFlytek, including 
GPT-3.5, ChatGPT, and GPT-4 from OpenAI, Claude from Anthropic, LLaMA-13B and Vicuna-13B (LLaMA based) from Meta, 
ChatGLM-6B, ChatGLM v2-6B, and ChatGLM-13B from Tsinghua, Ernie from Baidu, Qianwen from Alibaba, and Spark from iFlytek. 
The model size ranges from 6 billion to 175 billion.
The detailed information is listed in Table.~\ref{tab:llms}.

\begin{table}[h]
\centering
\caption{The Overview of Evaluated LLMs.}
\label{tab:llms}
\resizebox{0.97\textwidth}{!}{%
\begin{tabular}{@{}lrrrccc@{}}
\toprule
Model & Model Size & Training Data Composition & Temperature & Developer & Open Source & Access \\ \midrule
ChatGLM-6B & 6 billion & Chinese and English & 0.5 / 1 & Tsinghua & Yes & Local \\
ChatGLM v2-6B & 6 billion & Chinese and English & 0.5 / 1 & Tsinghua & YES & Local \\
ChatGLM-130B & 130 billion & Chinese and English & Not support & Tsinghua & No & Web \\
GPT-3.5 & 175 billion & Primarily English & 1.0 / 2 & OpenAI & No & API \\
ChatGPT & 175 billion & Primarily English & 1.0 / 2 & OpenAI & No & API \\
GPT-4 & Unknown & Primarily English & 1.0 / 2 & OpenAI & No & API \\
Claude & Unknown & Primarily English & Not support & Anthropic & No & Web \\
LLaMA-13B & 13 billion & Primarily English & 0.5 / 1 & Meta & Yes & Local \\
Vicuna-13B & 13 billion & Primarily English & 0.5 / 1 & UC Berkeley et al. & Yes & Local \\
Ernie & 175 billion & Chinese and English & Not support & Baidu & No & Web \\
Qianwen & Unknown & Chinese and English & Not support & Alibaba & No & Web \\
Spark & Unknown & Chinese and English & Not support & iFlytek & No & Web \\ \bottomrule
\end{tabular}%
}
\end{table}

\subsubsection{Evaluation Method}
We do not use prompt engineering like few-shot learning and chain-of-thought (CoT) to avoid unpredicted impacts, verified by related work~\cite{huang2023ceval}.
They find the CoT performs better on some models (e.g., ChatGLM-6B, +3\%) and worse on others (e.g., Chinese-LLaMA-13B, -11.9\%).
We also conduct an experiment to evaluate the impact of few-shot learning. On ChatGLM v2-6B, we randomly select several question-and-answer pairs as examples to evaluate the impact. Then we use different random seeds and run them three times for each seed. Table.~\ref{tab:few-shot} shows the result. The bold text shows the best results (about 37.54\%), and the underline text shows the worst results (about 32.69\%). The accuracy gap using different random seeds is large, achieving about 5\%. 

\begin{table}[h]
\centering
\footnotesize
\caption{Few-shot experiment on ChatGLM v2-6B using different random seeds and run three times for each seed. Bold text represents the best results, and underlined test represents the worst results.}
\label{tab:few-shot}
\begin{tabular}{@{}llll@{}}
\toprule
Seed & Run \#1 & Run \#2 & Run \#3 \\ \midrule
0 & \textbf{37.54\%} & \textbf{37.54\%} & \textbf{37.32\%} \\
1 & 33.44\% & 33.33\% & 34.52\% \\
2 & {\ul 32.15\%} & {\ul 34.84\%} & {\ul 32.69\%} \\
3 & 35.28\% & 35.49\% & 34.41\% \\
4 & 33.23\% & 33.01\% & 33.33\% \\ \bottomrule
\end{tabular}%
\end{table}

\subsection{Experiment Setup}~\label{evasetup}


We locally deploy ChatGLM-6B and ChatGLM v2-6B on a single NVidia V100 GPU and deploy LLaMA-13B and Vicuna-13B on four NVidia 1080 Ti GPUs. For close-sourced models that provide API access, we perform evaluations using their APIs, including GPT-3.5, ChatGPT, and GPT-4. For other close-sourced models which do not offer API access but provide web-based products, we perform evaluations by simulating user input in the browser, including Claude, ChatGLM-130B, Ernie, Qianwen, and Spark. 
To ensure fairness, we set the temperature parameter as half of the maximum value of the model for all twelve models, except for the web-based products, which do not provide the interface to set the temperature parameter. 

\begin{figure}[h]
    \centering
    \includegraphics[width=0.8\textwidth]{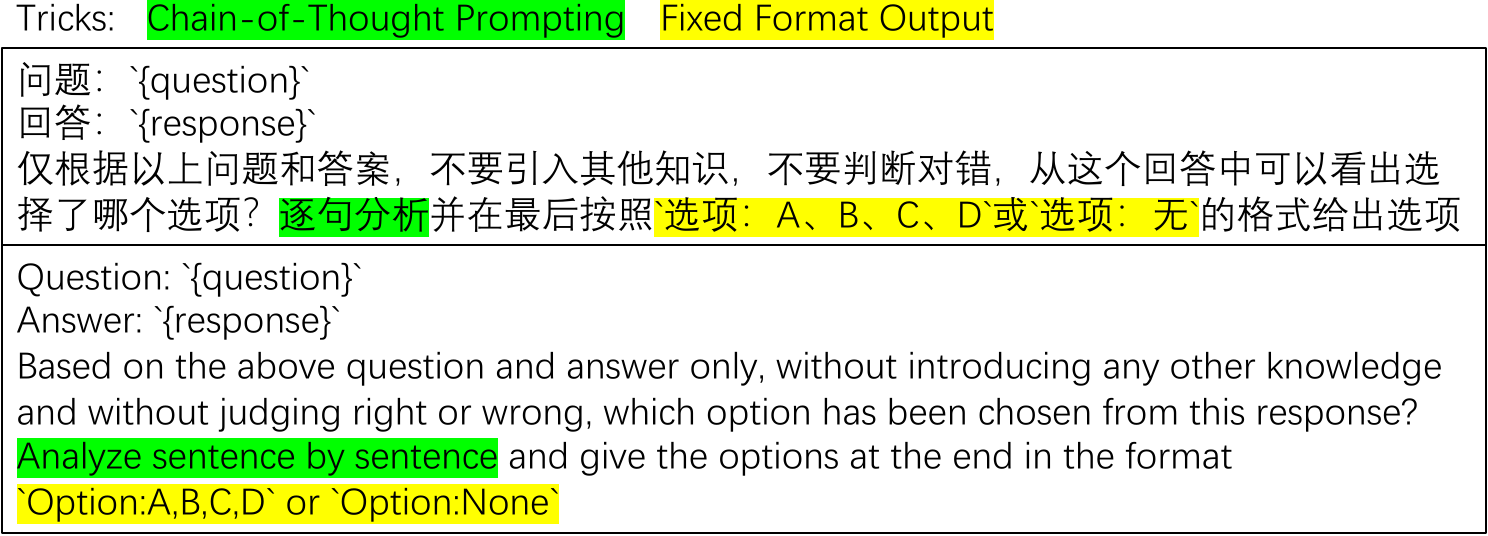}
    \caption{The Prompt Used to Extract and Judge Answers.}
    \label{fig:gpt-4_prompt}
\end{figure}

For the response extraction and judgment, we use the HRE algorithm illustrated in Algorithm~\ref{alg:pattern_sets} and GPT-4 to avoid the unpredictable performance impacts of prompt engineering. We set N as 10, M as 100, and minimum\_limit as 2, 
 which means repeating 10 iterations and randomly selecting 100 responses during each iteration. If a response format occurs at least two times, we will add this pattern.
We collect 66744 responses from LLMs, and the regex patterns identified by the HRE algorithm are shown in Table~\ref{tab:patterns}.
These patterns support extracting and judging 67\% answers of the total responses,  demonstrating the effectiveness of our algorithm. 
The remaining 33\% responses that have no unified regex pattern are processed by GPT-4, and the judgments are manually checked.
We use the chain-of-thought (CoT) prompt shown in Fig.~\ref{fig:gpt-4_prompt} to fix the judgment output of GPT-4 when extracting the answers. Note that this prompt is used to judge the correctness of the remaining responses without impacting the model evaluation process. 
\begin{table}[h]
\centering
\caption{Regex Patterns Identified by HRE Algorithm.}
\label{tab:patterns}
\resizebox{0.97\textwidth}{!}{%
\begin{tabular}{@{}lrr@{}}
\toprule
Regex & Number & Ratio \\ \midrule
Not Matching & 32875 & 33\% \\
【?答案】?(?:和原因)?(?:为|(?:应该)?是|选择)?{[}:：{]}?\textbackslash{}s?(?:选项)?({[}A-Z{]}){[}\textasciicircum{}A-Z{]}*?(?:。|\$) & 29508 & 29\% \\
\textasciicircum{}(?:选项\textbackslash{}s?)?({[}A-Z{]}){[}\textbackslash{}.。，\textbackslash{}s{]}{[}\textasciicircum{}A-Z{]}*?\$ & 20038 & 20\% \\
选择?：?\textbackslash{}s?({[}A-Z{]})(?:选项)?{[}\textasciicircum{}A-Z{]}*?(?:。|\$) & 6179 & 6\% \\
(?:{[}Aa{]}nswer|statement|{[}Ss{]}olution)\textbackslash{}s?(?:to   (?:your|this) question)? is:? (?: option )?\textbackslash{}"?({[}A-Z{]})\textbackslash{}"?(?:\textbackslash{}.|。|:|\$) & 3706 & 4\% \\
作为一个人工智能语言模型，我还没学习如何回答这个问题，您可以向我问一些其它的问题，我会尽力帮您解决的。 & 2299 & 2\% \\
\textasciicircum{}(?:答案{[}：:{]}\textbackslash{}s?)?({[}A-Z{]})(?:\textbackslash{}.|。|\textbackslash{}s)?\$ & 2173 & 2\% \\
\textasciicircum{}(?:选项\textbackslash{}s?)?({[}A-Z{]}){[}\textbackslash{}.。，\textbackslash{}s{]}(?:{[}\textasciicircum{}A-Z{]}|(\textbackslash{}1))*?\$ & 1765 & 2\% \\
(?:选项)?\textbackslash{}s?({[}A-Z{]})\textbackslash{}s?{[}是|为{]}正确答案 & 683 & 1\% \\
\textasciicircum{}(?:答案{[}：:{]}\textbackslash{}s?)?((?:{[}A-Z{]}\textbackslash{}s?{[}、\textbackslash{}s\textbackslash{}.-{]}?\textbackslash{}s?)+)。?\$ & 458 & 0\% \\
\textasciicircum{}\textbackslash{}s*\$ & 432 & 0\% \\ \bottomrule
\end{tabular}%
}
\end{table}


\subsection{Evaluation Results}~\label{evaresult}

We report the evaluation results including multi-granularity, multimodal, and human-referenced difficulty level, and response repeatability benchmarking. Further, using comprehensive metrics, we evaluate the accuracy of LLMs under different cases.

\subsubsection{Multi-granularity Benchmarking}

Table~\ref{tab:accuracy} shows the multi-granularity benchmarking results at a per-dataset, per-ability branch, and per-knowledge granularities. 
Note that the ``Human" column indicates human accuracy as a reference.
The average accuracy on the whole dataset is 60.91\%.
GPT-4 outperforms others largely on many ability branches and knowledge categories. 
Most LLMs have better understanding abilities than common sense and reasoning abilities, and the reasoning ability performs the worst.
Compared to the human accuracy, GPT-4 outperforms humans by 15.84\% for the common sense branch and has similar accuracy to humans (73.16\% vs. 74.82\%) for the understanding branch, while performing worse than humans for the reasoning branch with a gap of 25.94\%.


\begin{table}[h]
\centering
\caption{Multi-granularity Benchmarking Results.}
\label{tab:accuracy}
\resizebox{\textwidth}{!}{%
\begin{tabular}{@{}lllllllllllllll@{}}
\toprule
Ability & Knowledge & ChatGLM & ChatGLM v2-6B & ChatGLM-130B & GPT-3.5 & ChatGPT & GPT-4 & Claude & LLaMA-13B & Vicuna-13B & Ernie & Qianwen & Spark & {\ul Human} \\ \midrule
\multirow{7}{*}{Common   Sense} & Avg & 32.15\% & 46.30\% & 36.99\% & 34.88\% & 40.28\% & \textbf{65.84\%} & 36.47\% & 23.25\% & 23.97\% & 31.28\% & 17.23\% & 19.65\% & {\ul 50.00\%} \\ \cmidrule(l){2-15} 
 & Humanities & 31.67\% & 41.67\% & 28.06\% & 19.17\% & 36.94\% & \textbf{56.11\%} & 34.44\% & 18.33\% & 18.89\% & 25.56\% & 17.22\% & 9.72\% & {\ul 48.54\%} \\ \cmidrule(l){2-15} 
 & Technology & 33.78\% & 47.11\% & 46.00\% & 36.89\% & 48.00\% & \textbf{75.78\%} & 36.22\% & 28.44\% & 26.22\% & 35.56\% & 22.22\% & 37.78\% & {\ul 51.10\%} \\ \cmidrule(l){2-15} 
 & Law & 29.25\% & 44.16\% & 31.50\% & 35.30\% & 33.61\% & \textbf{57.81\%} & 33.47\% & 22.22\% & 25.74\% & 30.52\% & 13.50\% & 8.72\% & {\ul 49.00\%} \\ \cmidrule(l){2-15} 
 & Geography & 36.63\% & 52.67\% & 44.03\% & 32.10\% & 43.21\% & \textbf{79.42\%} & 34.98\% & 21.40\% & 20.99\% & 31.28\% & 23.87\% & 35.39\% & {\ul 49.43\%} \\ \cmidrule(l){2-15} 
 & Politics & 26.67\% & 65.56\% & 41.11\% & 82.22\% & 54.44\% & \textbf{68.89\%} & 56.67\% & 22.22\% & 18.89\% & 35.56\% & 6.67\% & 14.44\% & {\ul 60.70\%} \\ \cmidrule(l){2-15} 
 & Economics & 42.22\% & 41.11\% & 47.78\% & 44.44\% & 45.56\% & \textbf{78.89\%} & 53.33\% & 31.11\% & 32.22\% & 34.44\% & 14.44\% & 17.78\% & {\ul 49.10\%} \\ \midrule
\multirow{11}{*}{Reasoning} & Avg & 27.95\% & 29.29\% & 29.29\% & 25.82\% & 32.95\% & \textbf{36.03\%} & 27.87\% & 18.85\% & 21.75\% & 24.91\% & 23.91\% & 30.75\% & {\ul 61.97\%} \\ \cmidrule(l){2-15} 
 & Graphic Reasoning & \textbf{26.60\%} & 24.41\% & 21.89\% & 12.96\% & 21.21\% & 9.43\% & 27.27\% & 21.21\% & 18.35\% & 20.37\% & 20.37\% & 20.20\% & {\ul 71.76\%} \\ \cmidrule(l){2-15} 
 & Definition Judgment & 32.22\% & 34.44\% & 57.78\% & 40.56\% & 63.33\% & \textbf{72.22\%} & 36.11\% & 31.11\% & 24.44\% & 33.89\% & 20.00\% & 41.11\% & {\ul 76.20\%} \\ \cmidrule(l){2-15} 
 & Analogical Reasoning & 26.10\% & 31.86\% & 37.88\% & 12.72\% & \textbf{38.02\%} & 29.72\% & 22.09\% & 12.32\% & 17.67\% & 25.84\% & 14.59\% & 34.94\% & {\ul 67.08\%} \\ \cmidrule(l){2-15} 
 & Logical Judgment & 34.48\% & 38.89\% & 44.12\% & 37.58\% & 45.92\% & \textbf{61.27\%} & 42.65\% & 21.90\% & 32.03\% & 35.46\% & 31.21\% & 38.07\% & {\ul 70.73\%} \\ \cmidrule(l){2-15} 
 & Mathematical Calculation & 27.04\% & 26.99\% & 24.77\% & 28.61\% & 30.00\% & \textbf{36.11\%} & 25.42\% & 20.28\% & 20.32\% & 25.28\% & 27.27\% & 34.81\% & {\ul 49.53\%} \\ \cmidrule(l){2-15} 
 & Numerical Reasoning & 22.22\% & 25.93\% & 27.05\% & 26.57\% & 32.37\% & \textbf{34.62\%} & 28.34\% & 15.46\% & 23.03\% & 17.71\% & 21.74\% & 24.15\% & {\ul 67.08\%} \\ \cmidrule(l){2-15} 
 & Text Analysis & \textbf{38.89\%} & 52.22\% & 0.00\% & 36.67\% & \textbf{38.89\%} & 34.44\% & 22.22\% & 11.11\% & 21.11\% & 28.89\% & 18.89\% & 0.00\% & {\ul 72.10\%} \\ \cmidrule(l){2-15} 
 & Table Analysis & \textbf{53.33\%} & 25.56\% & 22.22\% & 25.56\% & 17.78\% & 46.67\% & 23.33\% & 21.11\% & 27.78\% & 6.67\% & 30.00\% & 12.22\% & {\ul 76.36\%} \\ \cmidrule(l){2-15} 
 & Graphic Analysis & 20.00\% & 22.22\% & 20.00\% & 20.00\% & 13.33\% & 16.67\% & 20.00\% & 13.33\% & 14.44\% & 27.78\% & \textbf{30.00\%} & 10.00\% & {\ul 78.50\%} \\ \cmidrule(l){2-15} 
 & Comprehensive Analysis & 32.22\% & 31.11\% & 18.89\% & 33.33\% & 23.33\% & \textbf{37.78\%} & 36.67\% & 12.22\% & 30.00\% & 10.00\% & 10.00\% & 13.33\% & {\ul 59.00\%} \\ \midrule
\multirow{5}{*}{Understanding} & Avg & 44.53\% & 57.42\% & 46.22\% & 44.89\% & 62.58\% & \textbf{73.16\%} & 42.84\% & 26.40\% & 33.78\% & 39.82\% & 22.04\% & 40.27\% & {\ul 74.82\%} \\ \cmidrule(l){2-15} 
 & Passage Reading & 51.11\% & 68.40\% & 60.25\% & 55.80\% & 73.83\% & \textbf{79.51\%} & 48.40\% & 29.88\% & 43.70\% & 43.70\% & 24.44\% & 51.85\% & {\ul 74.66\%} \\ \cmidrule(l){2-15} 
 & Sentence Grammar & 41.94\% & 45.56\% & 43.33\% & 35.00\% & 56.67\% & \textbf{66.11\%} & 38.33\% & 26.39\% & 28.33\% & 40.28\% & 22.78\% & 41.39\% & {\ul 74.43\%} \\ \cmidrule(l){2-15} 
 & Fill-in-the-blank & 34.44\% & 51.48\% & 42.22\% & 33.33\% & 44.44\% & \textbf{67.41\%} & 41.85\% & 25.93\% & 27.04\% & 39.63\% & 24.81\% & 32.22\% & {\ul 78.38\%} \\ \cmidrule(l){2-15} 
 & Long Text Reading & 55.56\% & 73.33\% & 6.67\% & 70.00\% & \textbf{90.00\%} & \textbf{90.00\%} & 38.89\% & 12.22\% & 31.11\% & 21.11\% & 0.00\% & 7.78\% & {\ul 66.40\%} \\ \bottomrule
\end{tabular}%
}
\end{table}

\begin{figure}[h]
    \centering
    \includegraphics[width=0.97\textwidth]{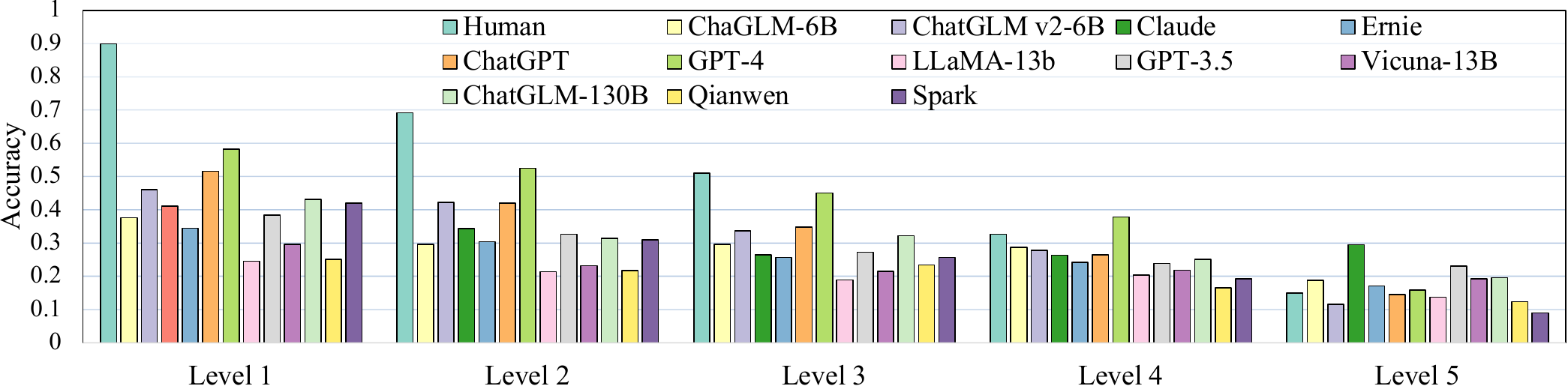}
    \caption{The Average Accuracy on Five Difficulty Levels. Level 1 is the easiest, and Lever 5 is the most difficult. The human label means the human accuracy on that level.}
    \label{fig:accuracy_levels}
\end{figure}

\begin{figure}[h]
    \centering
    \includegraphics[width=0.97\textwidth]{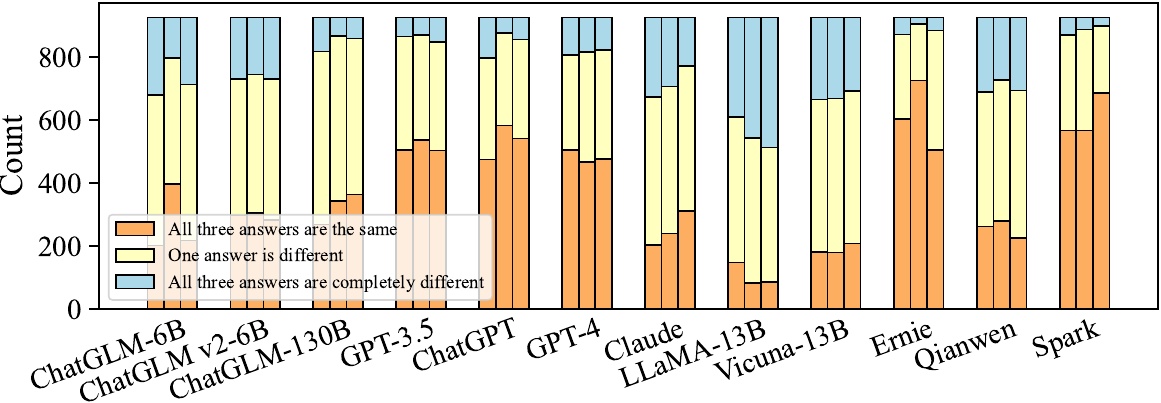}
    \caption{Sensibility to Prompts and Response Repeatability of LLMs.}
    \label{fig:answer_stability}

\end{figure}

\begin{figure}[ht]
    \centering
    \includegraphics[width=0.97\textwidth]{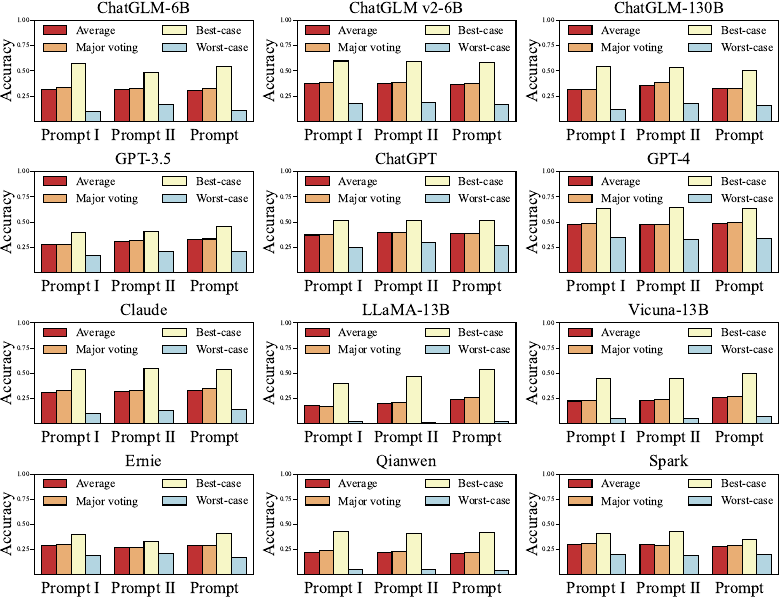}
    \caption{The Average, Worst-case, Best-case, Majority Voting Accuracy of LLMs.}
    \label{fig:answer_accuracies}
\end{figure}

\subsubsection{Multimodal Benchmarking}



We mainly perform the multimodal benchmarking on Ernie, ChatGLM-130B, Qianwen, and Spark models, since they provide Internet connectivity and can access images. 
For other eight models that do not provide Internet connectivity including GPT-3.5, ChatGPT, GPT-4, Claude, LLaMA-13B, Vicuna-13B, ChatGLM-6B, and ChatGLM v2-6B, we also input the image url to them. 

For the eight models that have no image processing ability, the ideal response is to admit they have no such ability and cannot comprehend images. However, our evaluations find that only slight responses of GPT-3.5, ChatGPT, and GPT-4 can admit the image processing limitations. The other models just generate hallucinational and nonsense responses.

For the four models that have image processing ability, through comprehensive multimodal benchmarking, we discover that none of the evaluated LLMs accurately comprehend image content. When facing a multimodal question that contains both text and image, their responses are either text or text with images. However, we find that the output response has low correlation with the input image. Even though Ernie has a certain probability to accurately comprehend the conventional images like ImageNet~\cite{deng2009imagenet}, however, the accuracy decreases largely (nearly zero) when facing geometric images.





\subsubsection{Difficulty Benchmarking using Human-referenced Accuracy}

We evaluate the ability to solve questions with different difficulty levels, using human accuracy as references, as shown in Fig.~\ref{fig:accuracy_levels}. Note that Level 1 to Level 5 is from simple to complex, with the easiest Level 1 and the most difficult Lever 5. The human label indicates the human accuracy on that level, and humans can achieve 89.92\%, 69.20\%, 51.11\%, 32.64\%, and 14.98\% average accuracy from Level 1 to 5, respectively. 

From Level 1 to 4, GPT-4 performs best among all twelve LLMs. For Level 5,  Claude performs the best.
Another exciting and counterintuitive phenomenon is that humans usually perform better than LLMs for simple difficulty levels like Levels 1 to 3. In contrast, humans perform worse than some LLMs for challenging levels like Levels 4 and 5. For example, GPT-4 has higher average accuracy than humans on Level 4. On Level 5, many LLMs have similar or even higher average accuracy than humans, including ChaGLM, Claude, Ernie, ChatGPT, GPT-4, GPT-3.5, Vicuna-13b, and ChatGLM-130b. The gap between the human and the best LLM on each level is 31.67\%, 16.68\%, 6.09\%, -5.23\%, -14.51\%, respectively. 


\subsubsection{Response Repeatability}~\label{Repeatability}
We further evaluate the repeatability of the responses. For each question, we ask LLMs using three different prompt types and each of which repeats three times. The three prompt types are (i) only question without any prompt; add (ii) ``The answer is:" and (iii) ``The answer and the reason are:" at the end of the question, respectively. Fig.~\ref{fig:answer_stability} shows the results. We classify the responses into three categories: (1) all three answers are the same, like choosing A three times, (2) one answer is different, like A, A, and B for three answers. And (3) all three answers are entirely different, like A, B, and C for three answers. Note that the (1) category means the best repeatability while (3) means the worst repeatability. We find that Spark and Ernie have the best response repeatability, while LLaMA-13B is the worst.

\subsubsection{The Average, Worst-case, Best-case, Majority Voting Accuracy}


Fig.~\ref{fig:answer_accuracies} presents the accuracy results under the average, worst, best, and majority voting cases. We also use three kinds of prompt types, which use the same prompt setting with the above response repeatability evaluation.
We find the following observations. (1) GPT-4 achieves the highest accuracy under four cases. (2) ChatGLM v2-6B performs better than ChatGLM-130B, which means the model architecture and the quality of training data are more important than merely increasing the model size. (3) The worst-case accuracy of LLMs is significantly below corresponding average one, which means the model not always gives a correct answer during three times evaluation, indicating a poor reliability of LLMs. (4) The best-case accuracy of LLMs is much higher than the other cases, which means the model has a high probability to give a correct answer during three times evaluation. (5) The majority voting accuracy is similar with the average one, which means most of the time, the model can give a correct answer.

\section{Conclusion}

This paper provides a multi-granularity, multimodal, human-referenced, and auto-scoring benchmark for evaluating large language models -- AGIBench, including a question dataset, auto-scoring evaluation, and comprehensive metrics. Through labeling each question with four attributes, including ability branch, knowledge, difficulty, modal, AGIBench supporting multi-granularity benchmarking at per-dataset, per-ability branch, per-knowledge, per-difficulty level, per-modal, and per-question granularities. We use the accuracy rate answered by millions of well-educated humans to label each question's difficulty level and include text and image modals. We further propose an HRE algorithm to avoid the unpredictable impacts of prompt engineering. Instead of only using average accuracy as a metric, we define multi-dimensional metrics to evaluate the LLMs comprehensively. Our experiments on twelve LLMs show the effectiveness of AGIBench.

%
%
%

\end{CJK}
\end{document}